%% file: main.tex
\documentclass{article} 
\usepackage{iclr2025_conference,times}
\usepackage{subcaption}

\usepackage{listings}
\usepackage[normalem]{ulem}
\usepackage{microtype}
\usepackage{amsmath}
\usepackage{amssymb}
\usepackage{mathtools}
\usepackage{amsthm}
\usepackage{xspace}
\usepackage{hyperref}
\usepackage[capitalize,noabbrev]{cleveref}

\usepackage{natbib}
\PassOptionsToPackage{table}{xcolor}

\usepackage{url}
\usepackage{booktabs}
\usepackage{graphicx}
\usepackage[ruled,vlined]{algorithm2e}
\usepackage{algorithmic}
\usepackage{tikz}
\usetikzlibrary{tikzmark,calc}
\usepackage{multirow}
\usepackage{xcolor}
\usepackage{colortbl}
\usepackage{pifont} 
\usepackage{makecell}
\definecolor{lightblue}{rgb}{0.1, 0.1, 0.9} 
\usepackage{fancyhdr}

\theoremstyle{plain}

\theoremstyle{definition}

\theoremstyle{remark}

\renewcommand{\headrulewidth}{0pt}
\title{SageBwd: A Trainable Low-bit Attention}

\author{
    \hspace{-.2em}Jintao Zhang\thanks{Equal contribution.},~ 
    Marco Chen\footnotemark[1],~ 
    Haoxu Wang\footnotemark[1],~
    Kai Jiang,~
    Ion Stoica,~ 
    Joseph E. Gonzalez,~ \\
    \textbf{Jianfei Chen,}~ 
    \textbf{Jun Zhu} \\
    ~Tsinghua University, UC Berkeley\\
    \texttt{\{zhang-jt24@mails., jianfeic@, dcszj@\}tsinghua.edu.cn} 
}

\newcommand{\vQ}{\mathbf{Q}}
\newcommand{\vK}{\mathbf{K}}
\newcommand{\vV}{\mathbf{V}}
\newcommand{\vdQ}{\mathbf{dQ}}
\newcommand{\vdK}{\mathbf{dK}}
\newcommand{\vdV}{\mathbf{dV}}
\newcommand{\vS}{\mathbf{S}}
\newcommand{\vdS}{\mathbf{dS}}
\newcommand{\vP}{\mathbf{P}}
\newcommand{\vdP}{\mathbf{dP}}

\newcommand{\vO}{\mathbf{O}}
\newcommand{\vdO}{\mathbf{dO}}

\newcommand{\our}{\texttt{SageBwd}\xspace}

\newcommand{\jt}[1]{\textcolor{blue}{{#1}}\xspace}

\newcommand{\annotate}[1]{\textcolor{gray}{{#1}}\xspace}

\definecolor{deepgreen}{rgb}{0.0, 0.5, 0.0}  
\definecolor{deepred}{rgb}{0.6, 0.0, 0.0}

\iclrfinalcopy 
\begin{document}

\maketitle

\begin{abstract}
Low-bit attention, such as SageAttention, has emerged as an effective approach for accelerating model inference, but its applicability to training remains poorly understood. In prior work, we introduced SageBwd, a trainable INT8 attention that quantizes six of seven attention matrix multiplications while preserving fine-tuning performance. However, SageBwd exhibited a persistent performance gap to full-precision attention (FPA) during pre-training. In this work, we investigate why this gap occurs and demonstrate that SageBwd matches full-precision attention during pretraining. Through experiments and theoretical analysis, we reach a few important insights and conclusions: \textbf{(i)} QK-norm is necessary for stable training at large tokens per step, \textbf{(ii)} quantization errors primarily arise from the backward-pass score gradient $\mathbf{dS}$, \textbf{(iii)} reducing tokens per step enables SageBwd to match FPA performance in pre-training, and \textbf{(iv)} K-smoothing remains essential for training stability, while Q-smoothing provides limited benefit during pre-training. 
\end{abstract}

\input{src/1-Introduction}

\input{src/2-Related_work}

\input{src/3-Preliminary}

\input{src/4-Method}
\input{src/5-Experiment}

\input{src/6-Conclusion}

\bibliography{main}
\bibliographystyle{iclr2025_conference}

\newpage

\appendix
\input{src/appendix.tex}

\end{document}

%% file: src/1-Introduction.tex
\section{Introduction}
\label{sec:intro}

\paragraph{Motivation.}
The efficiency of attention~\citep{vaswani2017attention} is critical for modern generative models, particularly
as context lengths continue to grow and the quadratic complexity of scaled
dot-product attention becomes a bottleneck~\citep{vaswani2017attention, jiang2024minference}.
Low-bit quantization offers a promising approach to reducing this cost by enabling
the use of low-precision Tensor Cores on GPUs~\citep{chen2020statistical}.
Recent methods such as SageAttention~\citep{zhang2025sageattention,zhang2024sageattention2,zhang2025sageattention2++}
and FlashAttention3~\citep{shah2024flashattention} have shown that low-bit attention
can be highly effective for inference; however, its applicability to training,
particularly large-scale pre-training, remains less well understood.

\nocite{hu2025quant,zhang2025int8train,zhang2025sla,zhang2026sla2,zhangspargeattention,zhangspargeattention2,zhang2025turbodiffusion,zhangefficient,yang2025sparse,hu2026residual,xi2026quant,xiang2026geometry,jiang2025cascadia,jiang2025hexgen3}

\paragraph{Challenge.}
Designing a trainable low-bit attention mechanism is challenging because the backward
pass is substantially more sensitive to quantization error than the forward pass.
In particular, computing gradients involves products of small-magnitude tensors and repeated error propagation
through the chain rule, which can amplify quantization errors. Moreover, quantization error in the forward output $\mathbf{O}$ propagates
directly through the backward computation, inducing deviations even when
the backward pass matrix multiplications (MatMuls) themselves are executed in higher precision. 

\paragraph{Contributions and insights.}
In prior work~\citep{zhang2025sageattention3}, we introduced \our, a trainable low-bit
attention mechanism that quantizes six of the seven attention matrix multiplications to
INT8 while preserving fine-tuning performance. However, during pre-training, \our
exhibited a persistent performance gap relative to full-precision attention (FPA).
In this work, we provide theoretical analyses and make empirical observations regarding the sources of this gap and identify conditions under which
\our recovers FPA-level pre-training performance.

\textbf{Key findings.}
First, we identify the dominant source of training deviation as the $\mathbf{dS}$
tensor in the backward pass, whose small magnitude makes it particularly vulnerable
to upstream quantization error.
Second, we show that QK-norm stabilizes pre-training by constraining query--key
outliers.
Third, we find that reducing the number of tokens per optimization step allows \our
to match FPA pre-training performance, suggesting that increased gradient noise can
mitigate the impact of quantization error.
Finally, through targeted ablations, we show that K-smoothing remains necessary for
stable training, while Q-smoothing provides limited benefit in the pre-training
setting.

%% file: src/2-Related_work.tex
\section{Related Work}  \label{sec:related_work}
\paragraph{Hardware-efficient attention.}
A line of recent work accelerates attention by optimizing GPU kernel implementations.
FlashAttention~\citep{dao2022flashattention} reduces memory I/O by tiling attention
computation to on-chip SRAM, achieving significant speedups over standard attention.
FlashAttention2~\citep{dao2023flashattention} further improves parallelism and warp
partitioning, while FlashAttention3~\citep{shah2024flashattention} targets kernel-level
optimizations on Hopper GPUs.
Similarly, xFormers~\citep{xFormers2022} provides a collection of custom CUDA kernels
for efficient attention variants.

\paragraph{Low-bit and quantized attention.}
Another line of work accelerates attention by leveraging low-precision tensor cores.
SageAttention~\citep{zhang2025sageattention}, SageAttention2~\citep{zhang2024sageattention2},
and SageAttention2++~\citep{zhang2025sageattention2++} combine INT8 quantization with
outlier-smoothing techniques to enable efficient attention computation.
FlashAttention3~\citep{shah2024flashattention} proposes an FP8 attention variant;
however, it is not directly applicable to large generative models such as video
diffusion in a plug-and-play manner~\citep{zhang2024sageattention2}.
More broadly, these low-bit attention methods are primarily designed for inference
and do not support training, limiting their applicability in pre-training
and fine-tuning settings.

\paragraph{Trainable low-bit attention.}
SageAttention3~\citep{zhang2025sageattention3} introduces two complementary advances:
(i) an extension of SageAttention2++ that improves inference-side low-bit attention,
and (ii) \our, a trainable low-bit attention mechanism that quantizes most
attention matrix multiplications while preserving fine-tuning performance.
This work builds on the SageBwd component of SageAttention3 by analyzing the sources
of training instability in low-bit attention and characterizing the conditions
under which full-precision attention performance can be recovered
during pre-training.

%% file: src/3-Preliminary.tex
\section{Preliminaries}
\label{sec:preliminaries}

\paragraph{FlashAttention.}

Scaled dot-product attention computes $\mathbf{S} = \mathbf{Q}\mathbf{K}^\top,\mathbf{P} = \operatorname{softmax}(\mathbf{S}),\mathbf{O} = \mathbf{P}\mathbf{V}$ where $\mathbf{Q}, \mathbf{K}, \mathbf{V} \in \mathbb{R}^{N \times D}$ and
$\mathbf{P}, \mathbf{S} \in \mathbb{R}^{N \times N}$, with $N$ denoting the sequence length
and $D$ the head dimension. FlashAttention tiles the sequence dimension by chunking
$\mathbf{Q}, \mathbf{K}, \mathbf{V}$ into blocks
$\{\mathbf{Q}_i\}, \{\mathbf{K}_j\}, \{\mathbf{V}_j\}$, where
$\mathbf{Q}_i \in \mathbb{R}^{B_q \times D}$ and
$\mathbf{K}_j, \mathbf{V}_j \in \mathbb{R}^{B_{kv} \times D}$.
It then avoids the quadratic IO overhead of materializing $\mathbf{S}$ and
$\mathbf{P}$ in global memory by using an online softmax and fusing all
operations into a single kernel: $\mathbf{S}_{ij} = \mathbf{Q}_i \mathbf{K}_j^\top, \mathbf{P}_{ij} = \operatorname{OnlineSoftmax}(\mathbf{S}_{ij}),\mathbf{O}_i = \sum_j \mathbf{P}_{ij}\mathbf{V}_j$.

\paragraph{Quantization.}
Quantization accelerates matrix multiplication by representing high-precision
matrices with low-bit numeric formats and floating-point scale factors.
Given a high-precision matrix $\mathbf{X} \in \mathbb{R}^{m \times n}$, its INT8 quantization $\hat{\mathbf{X}}$ is defined as $
\hat{\mathbf{X}} := \operatorname{round}(\mathbf{X} / \delta_{\mathbf{X}}),
$
where $\delta_{\mathbf{X}} > 0$ is a scale factor, typically computed as
$\delta_{\mathbf{X}} = \max(|\mathbf{X}|)/127$, and stored in FP32.

Subsequently, given two FP16 matrices $\mathbf{A}$ and $\mathbf{B}$, their approximate matrix product under INT8 quantization is computed as 
\[
\mathbf{A}\mathbf{B} \approx \delta_{\mathbf{A}} \delta_{\mathbf{B}} \cdot
\hat{\mathbf{A}} \hat{\mathbf{B}},
\]
where the integer matrix multiplication $\hat{\mathbf{A}} \hat{\mathbf{B}}$ can be
accelerated using INT8 tensor cores.

The \emph{granularity} of quantization refers to the scope over which the scale factor
$\delta$ is computed.
Common choices include per-tensor, per-channel, and per-block quantization.
In \emph{per-block} quantization, a single scale factor is shared by all elements
within a block, e.g., a FlashAttention tile.

\paragraph{Q and K Smoothing in SageAttention.}

SageAttention~\citep{zhang2025sageattention,zhang2024sageattention2,zhang2025sageattention3} extends FlashAttention by quantizing $\mathbf{Q}$ and $\mathbf{K}$ to low precision for efficient inference. To mitigate the effect of channel-wise outliers prior to quantization, SageAttention introduced the preprocessing techniques of Q- and K-smoothing.
Given query and key blocks
$\mathbf{Q}_i \in \mathbb{R}^{B \times d}$ and $\mathbf{K}_j \in \mathbb{R}^{B \times d}$,
SageAttention computes a block-wise mean for queries and a global mean for keys:
\[
\boldsymbol{\mu}_{Q_i} = \operatorname{mean}_{\text{row}}(\mathbf{Q}_i), \qquad
\boldsymbol{\mu}_{K} = \operatorname{mean}_{\text{row}}(\mathbf{K}),
\]
where $\boldsymbol{\mu}_{Q_i}, \boldsymbol{\mu}_{K} \in \mathbb{R}^{1 \times d}$ and
$\mathbf{K}$ denotes the full key tensor.
The smoothed tensors are defined as
\[
\mathbf{Q}_i^{\mathrm{sm}} = \mathbf{Q}_i - \boldsymbol{\mu}_{Q_i}, \qquad
\mathbf{K}_j^{\mathrm{sm}} = \mathbf{K}_j - \boldsymbol{\mu}_{K}.
\]
The attention logits admit the decomposition
\[
\mathbf{Q}_i \mathbf{K}_j^\top
=
\mathbf{Q}_i^{\mathrm{sm}} \mathbf{K}_j^{\mathrm{sm}\,\top}
+ \boldsymbol{\mu}_{Q_i} \mathbf{K}_j^{\mathrm{sm}\,\top}
+ \mathbf{Q}_i^{\mathrm{sm}} \boldsymbol{\mu}_{K}^\top
+ \boldsymbol{\mu}_{Q_i} \boldsymbol{\mu}_{K}^\top.
\]
Since the softmax operation is invariant to adding a constant to each row,
SageAttention applies low-bit quantization to the smoothed tensors
$\mathbf{Q}_i^{\mathrm{sm}}$ and $\mathbf{K}_j^{\mathrm{sm}}$, computes the dominant
term $\mathbf{Q}_i^{\mathrm{sm}} \mathbf{K}_j^{\mathrm{sm}\,\top}$ using low-bit
tensor cores, and adds back the remaining low-rank bias term to recover the logits.
When only K-smoothing is applied, the additive bias term vanishes.

\paragraph{\our.}
SageAttention3~\citep{zhang2025sageattention3} proposes \our, a trainable
low-bit attention mechanism.
In the forward pass, \our applies K-smoothing prior to per-block INT8
quantization of $\mathbf{Q}\mathbf{K}^\top$, and uses a mixed per-token or
per-block quantization scheme for the $\tilde{\mathbf{P}}\mathbf{V}$ product.

In the backward pass, attention gradients involve the following matrix
multiplications:
\[
\mathbf{S} = \mathbf{Q}\mathbf{K}^\top,\quad
\mathbf{dV} = \mathbf{P}^\top \mathbf{dO},\quad
\mathbf{dP} = \mathbf{dO}\mathbf{V}^\top,\quad
\mathbf{dQ} = \mathbf{dS}\mathbf{K},\quad
\mathbf{dK} = \mathbf{dS}^\top \mathbf{Q},
\]
where
\[
\mathbf{dS} = \mathbf{P} \circ (\mathbf{dP} - \boldsymbol{\delta}\mathbf{1}^\top),
\qquad
\boldsymbol{\delta} = \operatorname{rowsum}(\mathbf{dO} \circ \mathbf{O}).
\]
\our retains $\mathbf{dP} = \mathbf{dO}\mathbf{V}^\top$ in FP16 precision,
while quantizing the remaining four matrix multiplications using per-block
INT8. This design choice avoids error amplification through $\mathbf{dS}$,
$\mathbf{dQ}$, and $\mathbf{dK}$ that arises when $\mathbf{dP}$ is quantized.
SageAttention3 demonstrates that this formulation preserves fine-tuning
performance, while its behavior during full pre-training remains less
well understood.
A pseudocode description is provided in \autoref{sec:append:algorithm}.

%% file: src/4-Method.tex
\section{Analysis of \our in Pretraining}
In this section, we analyze which design choices in \our are necessary to match full-precision attention (FPA) performance during pre-training.
Our analysis focuses on four central aspects:
(i) controlling query--key outliers via QK-norm,
(ii) identifying the most sensitive tensor in the INT8 backward pass,
(iii) understanding how tokens-per-step interacts with quantization noise,
and
(iv) characterizing the effect of the activation scale through controlled
QK standard deviation experiments.
Together, these analyses provide a mechanistic explanation for the observed behavior of \our.
In \autoref{sec:exp}, we empirically validate the resulting conclusions via pre-training experiments.

\subsection{Stabilizing Outliers with QK-Norm}
\label{subsec:qk}

\paragraph{QK-norm for logit stabilization.}
In scaled dot-product attention, the logits
$\mathbf{S} = \mathbf{Q}\mathbf{K}^\top / \sqrt{d}$ scale with the norms of
$\mathbf{Q}$ and $\mathbf{K}$.
During pre-training, these norms tend to increase, leading to large logits that
can saturate the softmax or trigger numerical instabilities, particularly under
low-precision arithmetic~\citep{logitControl_anson2025controllingchangesattentionlogits, dehghani2023scaling}.
QK-norm~\citep{qknorm_henry2020querykeynormalizationtransformers} addresses this issue
by applying RMS normalization to each token in $\mathbf{Q}$ and $\mathbf{K}$,
explicitly controlling their scale and keeping the logits within a numerically stable
range throughout training.

\paragraph{QK-norm for quantization.}
Beyond stabilizing the softmax, QK-norm is also useful in improving the
robustness of low-bit attention.
Prior work, such as SageAttention, combines channel-wise smoothing of $\mathbf{Q}$
and $\mathbf{K}$ with fine-grained quantization to mitigate the effect of extreme
outliers~\citep{zhang2025sageattention, zhang2024sageattention2}.
QK-norm complements these techniques: by compressing the dynamic range of
$\mathbf{Q}$ and $\mathbf{K}$, it reduces the effective quantization step size under
uniform INT8 quantization, pulling outliers closer to the rest of the distribution and improving quantization accuracy.
As shown in \autoref{subsec:exp:qknorm}, this effect is particularly important during
pre-training with \our.

\subsection{Sensitivity of $\mathbf{dS}$ in the Backward Pass}
\label{subsec:ds}

A central challenge in training low-bit attention is the accurate computation of
the softmax-gradient tensor $\mathbf{dS}$.
Empirically, \autoref{tab:error-analysis-components} shows that the discrepancy
between \our and FPA peaks at $\mathbf{dS}$, with errors further
propagating to $\mathbf{dQ}$ and $\mathbf{dK}$.
This behavior indicates that $\mathbf{dS}$ likely constitutes the primary numerical
bottleneck in the INT8 backward pass.

\paragraph{Why $\mathbf{dS}$ is intrinsically fragile.}
The sensitivity of $\mathbf{dS}$ stems from its systematically small magnitude.
Recall that
\[
\mathbf{dS} = \mathbf{P} \circ (\mathbf{dP} - \boldsymbol{\delta}),
\qquad
\boldsymbol{\delta} = \operatorname{rowsum}(\mathbf{dO} \circ \mathbf{O}),
\]
where $\mathbf{P}$ is the softmax output.
As shown in \autoref{sec:append:ds}, the RMS of $\mathbf{dS}$ admits the
upper bound
\[
\operatorname{RMS}(\mathbf{dS}) \le \frac{1}{\sqrt{N}}
\max_i \left\| \mathbf{dP}_i - \boldsymbol{\delta}_i \mathbf{1} \right\|_\infty,
\]
where $N$ is the sequence length.
This $1/\sqrt{N}$ scaling implies that $\mathbf{dS}$ becomes increasingly small
for long sequences, even when upstream gradients are well behaved.

\paragraph{Implications for INT8 quantization.}
INT8 quantization introduces approximately fixed absolute noise determined by the
quantization step size~\citep{jacob2018quantization}.
For tensors with large magnitude, this noise is often tolerable; however, when the
signal itself is small, the same absolute error translates into a large relative
error.
As a result, $\mathbf{dS}$ exhibits a much poorer effective signal-to-noise ratio
under INT8 quantization than other intermediate tensors.
This issue is exacerbated by the multiplicative structure of
$\mathbf{dS} = \mathbf{P} \circ (\mathbf{dP} - \boldsymbol{\delta})$, which combines
quantization noise from both forward-pass tensors ($\mathbf{P}$, $\mathbf{O}$) and
backward-pass tensors ($\mathbf{dP}$, $\mathbf{dO}$).

\paragraph{Empirical scale of $\mathbf{dS}$.}
We empirically verify this analysis by measuring the RMS values of $\mathbf{P}$,
$\mathbf{dP}$, and $\mathbf{dS}$ from a representative layer and head of a
QK-normed \our checkpoint that was trained over 78B tokens with 2.1M tokens per step and a sequence length of $N=4096$:
\[
\operatorname{RMS}(\mathbf{P}) \approx 5\times 10^{-3}, \quad
\operatorname{RMS}(\mathbf{dP}) \approx 5\times 10^{-5}, \quad
\operatorname{RMS}(\mathbf{dS}) \approx 1\times 10^{-7}.
\]
While the theoretical bound suggests that $\mathbf{dS}$ should be at most
$1/\sqrt{4096} \approx 1/64$ times the scale of $\mathbf{dP}$, we observe a ratio
closer to $500$ in practice.
Although the bound is loose, this discrepancy only further highlights how tightly
constrained the magnitude of $\mathbf{dS}$ is in realistic training settings.

\paragraph{Propagation to $\mathbf{dQ}$ and $\mathbf{dK}$.}
Finally, since $\mathbf{dQ}$ and $\mathbf{dK}$ are obtained through matrix multiplications
with $\mathbf{dS}$, quantization errors in $\mathbf{dS}$ propagate directly and are amplified by the norms of $\mathbf{Q}$ and $\mathbf{K}$. This amplification becomes more pronounced at longer sequence lengths, consistent
with observations in prior work~\citep{zhang2025sageattention3}.

\subsection{Effect of Tokens-per-Step}
\label{sec:analysis:tps}

We define tokens-per-step (TPS) as the total number of tokens processed in a
single optimizer update.
In our experiments (\autoref{sec:exp}), we fix the sequence length and vary the
global batch size, making TPS directly proportional to batch size.
We observe that TPS has a significant impact on pre-training behavior:
at a large TPS of 2.1M, \our consistently underperforms FPA,
whereas at a smaller TPS of 260K, \our matches FPA within noise.

\paragraph{Gradient noise and quantization error.}
Under a fixed token budget, increasing TPS results in fewer, more deterministic
gradient updates, while decreasing TPS yields more frequent updates with higher
stochastic gradient noise~\citep{smith2018dontdecaylearningrate}.
Prior work has shown that large-batch training reduces gradient noise and can
alter optimization dynamics~\citep{keskar2017largebatchtrainingdeeplearning}.
In the context of low-bit attention, we hypothesize that, at large TPS, the reduced
gradient noise makes systematic quantization error in the backward pass—particularly
along the sensitive $\mathbf{dS}$ path (\autoref{subsec:ds})—more salient to the optimizer.
This persistent, biased error may then influence the optimization trajectory,
leading to convergence toward a stable but suboptimal solution.

At smaller TPS, the inherent stochasticity of gradient updates is higher.
In this regime, INT8 quantization error likely acts as a small perturbation
relative to gradient noise and therefore may not significantly alter the training
trajectory, allowing \our to recover FPA-level pre-training performance. While this explanation provides a plausible mechanism linking TPS and quantization
error, we do not rule out the influence of other batch-size–dependent optimization effects.

\paragraph{Sequence length as a potential factor.}
In this work, we vary TPS only through batch size while holding the sequence length
fixed.
However, since $\mathbf{dQ}$ and $\mathbf{dK}$ involve matrix multiplications over
the sequence dimension $N$, longer sequences aggregate contributions from a larger
number of $\mathbf{dS}$ entries.
As a result, upstream errors may be further
amplified at larger sequence lengths.
A systematic study of how sequence length interacts with TPS and quantization error is left to future work.

\subsection{Effect of QK Standard Deviation on Quantization Error}
\label{subsec:qkstd}

To isolate the effect of activation scale on quantization error, we evaluate SageBwd
using synthetic Gaussian attention inputs in which the standard deviations of
$\mathbf{Q}$ and $\mathbf{K}$ ($\sigma_Q$, $\sigma_K$) are varied, while holding
$\sigma_V$ and $\sigma_{dO}$ fixed at 1.
This controlled setting removes optimizer dynamics and directly probes the
sensitivity of quantized attention to activation scale, simulating the typical growth of
query and key norms observed during pre-training.

As shown in Table~\ref{tab:error-analysis-qkstd}, the accuracy of \our degrades sharply
as $\sigma_Q$ and $\sigma_K$ increase.
While the output $\mathbf{O}$ and gradient $\mathbf{dV}$ remain relatively accurate,
the gradients $\mathbf{dQ}$ and $\mathbf{dK}$ exhibit severe error, with cosine
similarity dropping below 0.79 and relative $\ell_2$ error exceeding 0.66 at
$\sigma_{Q,K}=10$.

\begin{table}[ht]
  \centering
  \caption{Sage vs.\ FPA across random QKV with varying $\sigma_Q$ and $\sigma_K$}
  \label{tab:error-analysis-qkstd}
  \setlength{\tabcolsep}{4pt}
  \renewcommand{\arraystretch}{1.1}
  \begin{tabular}{c cc|cc|cc|cc}
    \toprule
    \multirow{2}{*}{$\sigma_{Q},\sigma_{K}$} &
      \multicolumn{2}{c}{Output} &
      \multicolumn{2}{c}{dQ} &
      \multicolumn{2}{c}{dK} &
      \multicolumn{2}{c}{dV} \\
    \cmidrule(lr){2-3} \cmidrule(lr){4-5} \cmidrule(lr){6-7} \cmidrule(lr){8-9}
     & CosSim & Rel-$\ell_2$ & CosSim & Rel-$\ell_2$ & CosSim & Rel-$\ell_2$ & CosSim & Rel-$\ell_2$ \\
    \midrule
     1  & 0.9999 & 0.0160 & 0.9998 & 0.0184 & 0.9998 & 0.0220 & 0.9999 & 0.0159 \\
     3  & 0.9992 & 0.0389 & 0.9971 & 0.0758 & 0.9970 & 0.0777 & 0.9992 & 0.0387 \\
     5  & 0.9982 & 0.0603 & 0.9798 & 0.2014 & 0.9799 & 0.2007 & 0.9982 & 0.0605 \\
     8  & 0.9953 & 0.0972 & 0.8900 & 0.4666 & 0.8886 & 0.4699 & 0.9953 & 0.0973 \\
    10  & 0.9933 & 0.1161 & 0.7823 & 0.6648 & 0.7820 & 0.6684 & 0.9933 & 0.1157 \\
    \bottomrule
  \end{tabular}
\end{table}

Intuitively, increasing $\sigma_Q$ and $\sigma_K$ inflates the dynamic range of tensors
involved in quantized matrix multiplications, increasing the quantization step size
under uniform INT8 quantization and thereby amplifying absolute quantization
noise~\citep{jacob2018quantization}.
These errors are especially harmful when they feed into the softmax gradient
computation, since $\mathbf{dS}$ has comparatively small magnitude
(\autoref{subsec:ds}), resulting in a poor effective signal-to-noise ratio.
Consequently, even moderate upstream absolute errors can translate into large relative
errors in $\mathbf{dS}$ and propagate to $\mathbf{dQ}$ and $\mathbf{dK}$.

This analysis also further clarifies the role of QK-norm in stabilizing pre-training, as mentioned in \autoref{subsec:qk}. Evidently, by normalizing $\mathbf{Q}$ and $\mathbf{K}$, QK-norm bounds their effective scale and
reduces the dynamic range seen by INT8 quantization, yielding much higher quantization accuracy.
However, since QK-norm includes a learned RMSNorm scale vector $\boldsymbol{\gamma}$, which
tends to increase gradually during pre-training~\citep{xiao2023smoothquant}, the effective $\sigma_Q$ and $\sigma_K$ may still grow over time.
Once this growth surpasses a critical error threshold, the quantization noise along the
$\mathbf{dS}$ path can become dominant, providing an explanation for the
pre-training instability observed at large tokens-per-step in \autoref{sec:exp}, even when QK-norm is applied.

%% file: src/5-Experiment.tex
\section{Experiments}  \label{sec:exp}
\paragraph{Core results.} At 260K tokens-per-step (TPS), \our pretrains with performance on par with full-precision attention (FPA), regardless of QK-norm. However, at 2.1M TPS, QK-norm is necessary to avoid loss explosion. In general, quantization error tends to spike at the intermediate tensor $\mathbf{dS}$. 

\subsection{Setup}
We implement \our using OpenAI Triton \citep{openaitriton} and conduct pretraining experiments with a 325M Llama model \citep{llama31model}) over 78B tokens of the OpenWebText dataset \citep{Gokaslan2019OpenWeb}. All runs use BF16 mixed precision and are trained on a single B200 or RTX4090 GPU node. Across all experiments, we use cosine learning rate scheduling, a context length of $4096$, a hidden dimension of $3072$, the GPT2 tokenizer, a norm epsilon of 1e-6, and a learning rate of 3e-5. By default, all experiments in this section apply K-smoothing but not Q-smoothing. 


\begin{figure}[t]
  \centering
  \begin{subfigure}{0.49\linewidth}
    \centering
    \includegraphics[width=\linewidth]{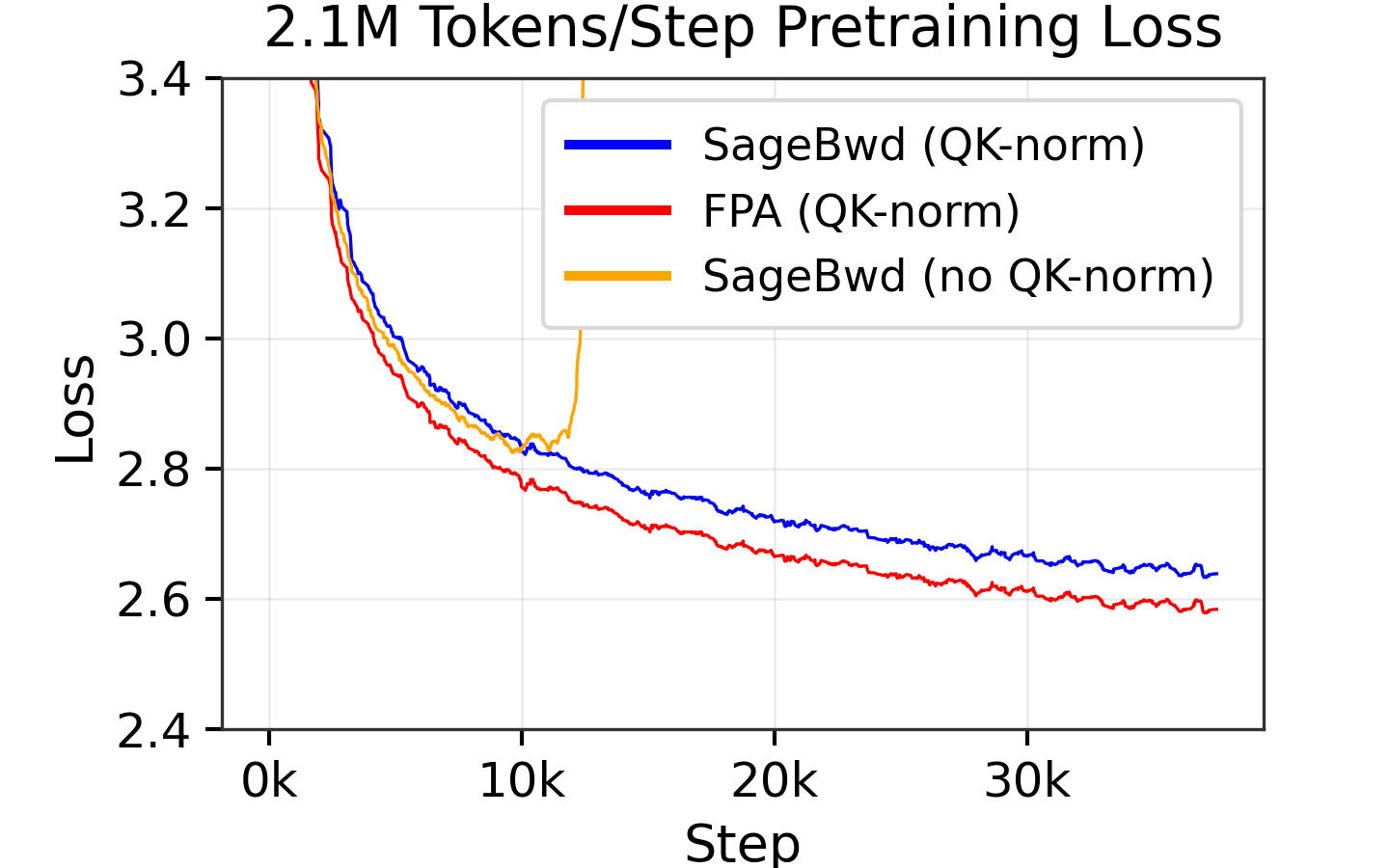}
    \caption{2.1M Tokens/Step}
    \label{fig:loss:a}
  \end{subfigure}\hfill
  \begin{subfigure}{0.49\linewidth}
    \centering
    \includegraphics[width=\linewidth]{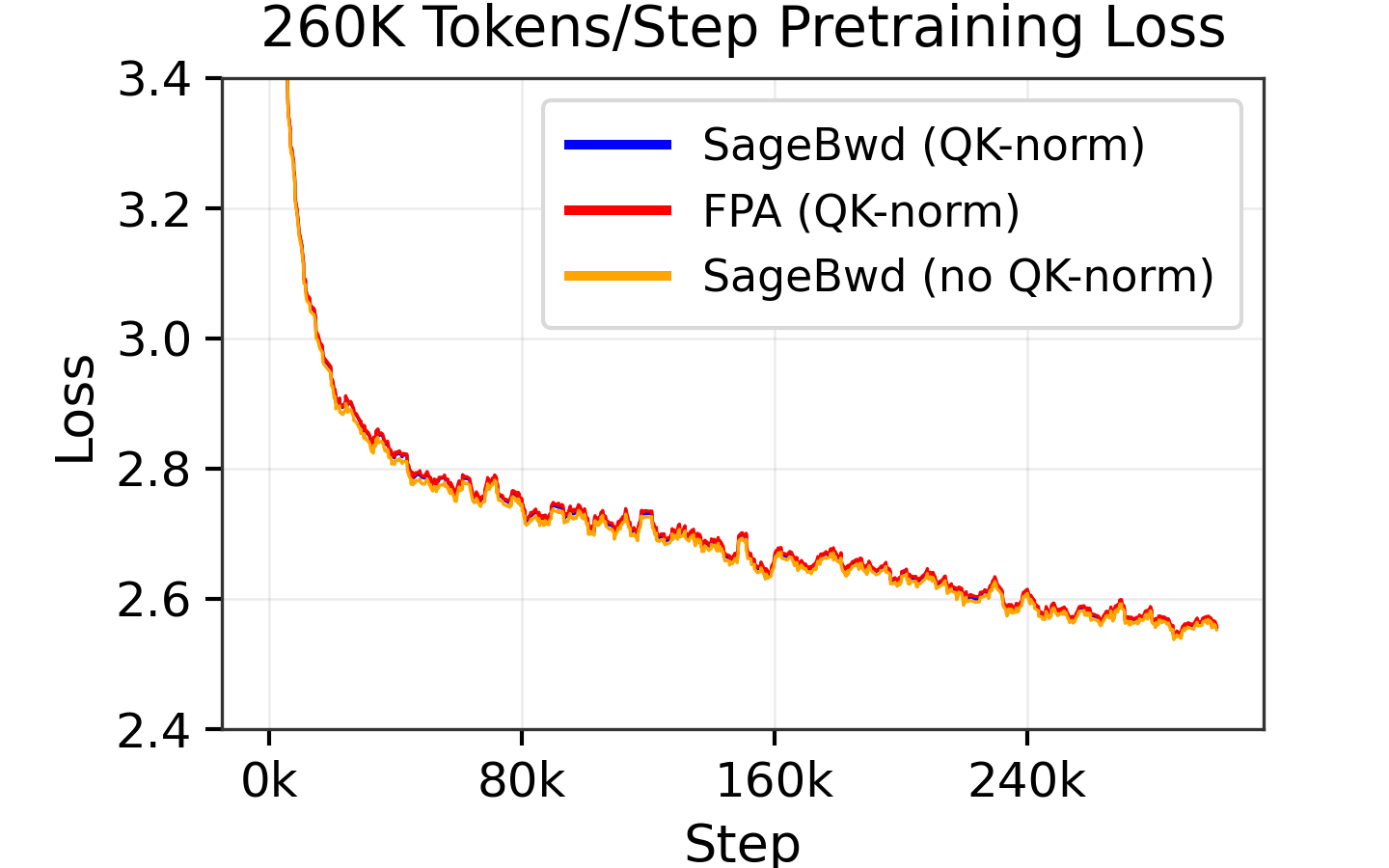}
    \caption{260K Tokens/Step}
    \label{fig:loss:b}
  \end{subfigure}
  \vspace{-.25em}
  \caption{Pretraining loss over 78B tokens under a different number of tokens/step}
  \label{fig:loss}
\end{figure}

\begin{table}[t]
  \centering
  \caption{Cosine similarity and relative $\ell_2$ error for intermediate tensors in \our (vs. FPA).}
  \label{tab:error-analysis-components}
  \setlength{\tabcolsep}{6pt}
  \renewcommand{\arraystretch}{1.15}
  \small
  \begin{tabular}{lcccc|cccc}
    \toprule
    Metric
      & $\boldsymbol{\delta}$ & $\mathbf{P}$ & $\mathbf{dP}$ & \textbf{$\mathbf{dS}$}
      & $\mathbf{O}$ & \textbf{$\mathbf{dQ}$} & \textbf{$\mathbf{dK}$} & $\mathbf{dV}$ \\
    \midrule
    CosSim
      & 0.9973 & 0.9917 & 1.0000 & \textbf{0.9789}
      & 0.9969 & \textbf{0.9664} & \textbf{0.9537} & 0.9985 \\
    Rel-L2
      & 0.0736 & 0.1293 & 0.0000 & \textbf{0.2045}
      & 0.0793 & \textbf{0.2579} & \textbf{0.3074} & 0.0540 \\
    \bottomrule
  \end{tabular}
  \vspace{-.25em}
\end{table}

\subsection{Effect of TPS on \our vs. full-precision attention pretraining}  
\autoref{fig:loss:a} and \autoref{fig:loss:b} compare the pre-training performance
of \our and FPA at 2.1M and 260K TPS, respectively.
At the larger TPS of 2.1M, \our exhibits a clear gap relative to FPA:
after 37.5k training steps with a global batch size of 512 (including 1k warmup
steps), \our reaches a loss of 2.640, whereas FPA attains 2.586.
In contrast, at the smaller TPS of 260K, where training is performed for 300k steps
with a global batch size of 64 (including 7.5k warmup steps), \our matches FPA
within noise, achieving a loss of 2.561 compared to 2.563 for FPA.

\subsection{QK-norm Is Necessary at High TPS}
\label{subsec:exp:qknorm}

As shown in \autoref{fig:loss:a}, at a large TPS of 2.1M, removing QK-norm leads to
training instability and eventual divergence.
This behavior is consistent with increased quantization error arising from
unconstrained query and key magnitudes.
In contrast, for the smaller-TPS runs in \autoref{fig:loss:b}, \our matches FPA
even without QK-norm.

Despite this apparent robustness at low TPS, intermediate-tensor analysis reveals
a different picture.
As reported in \autoref{sec:append:cossim_l2}, the non-normed runs exhibit
notably larger relative $\ell_2$ error and lower cosine similarity than their
QK-normed counterparts, even at 260K TPS.
This observation aligns with our hypothesis in \autoref{sec:analysis:tps} that
increased gradient noise at lower TPS can mask moderate quantization errors without
eliminating them.

Combined with the controlled activation-scale analysis in \autoref{subsec:qkstd},
these results indicate that QK-norm is a critical component for robust low-bit attention training at scale.

\subsection{Tracing intermediate tensor error}
In FlashAttention-style kernels, intermediate attention tensors such as
$\mathbf{P}$, $\mathbf{S}$, $\mathbf{dP}$, and $\mathbf{dS}$ are not explicitly
materialized, making direct accuracy inspection difficult.
To isolate quantization-induced error, we construct a pseudo-quantized FPA
baseline: we extract full-precision $\mathbf{Q}$, $\mathbf{K}$, $\mathbf{V}$,
and $\mathbf{dO}$ from layer 11 of \our with QK-norm in the 2.1M TPS run (the
most error-prone layer identified in \autoref{fig:metrics:cossim:a} and
\autoref{sec:append:cossim_l2}).
We then apply the \our INT8 quantize--dequantize scheme before each relevant
matrix multiplication in a PyTorch attention implementation and
compare all intermediate tensors against full-precision FPA using cosine similarity and relative $\ell_2$ error.

As shown in \autoref{tab:error-analysis-components}, most intermediates,
including $\mathbf{O}$ and $\mathbf{dV}$, remain very close to FPA.
In contrast, $\mathbf{dS}$ and its subsequent downstream gradients $\mathbf{dQ}$ and
$\mathbf{dK}$ exhibit substantially larger deviations.
This provides direct evidence that the $\mathbf{dS}$ computation constitutes
the primary quantization bottleneck in the backward pass of \our.
In this analysis, the upstream gradient $\mathbf{dO}$ is treated as
error-free; hence, $\mathbf{dP}$ appears perfectly accurate.

\subsection{Kernel Performance} 
\begin{figure}[h]
  \centering
  \includegraphics[width=\columnwidth]{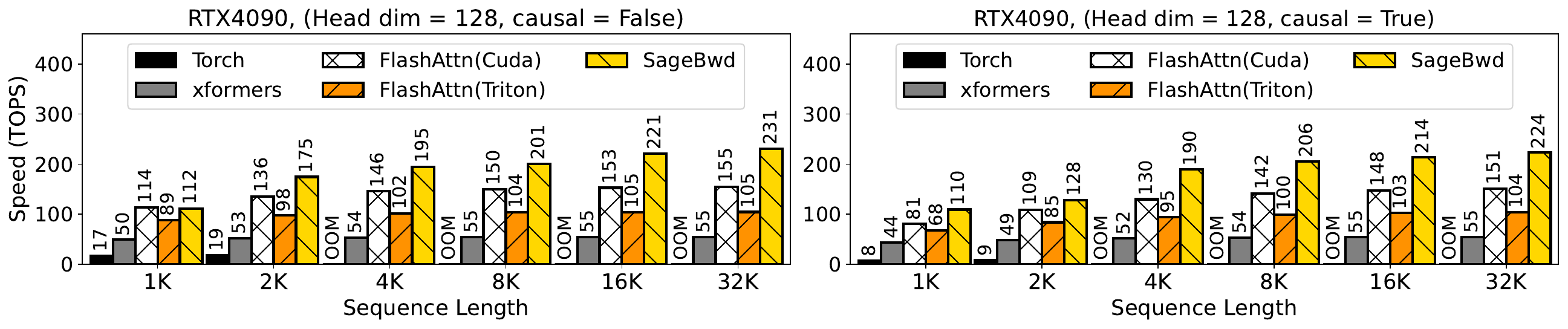} 
  \vspace{-.5em}
  \caption{Speed comparison between \texttt{SageBwd} and Baselines (\texttt{RTX4090}, headim=128).}
  \label{fig:sage_train_h128_speed} 
\end{figure}

\begin{figure}[h]
    \centering
    \includegraphics[width=\columnwidth]{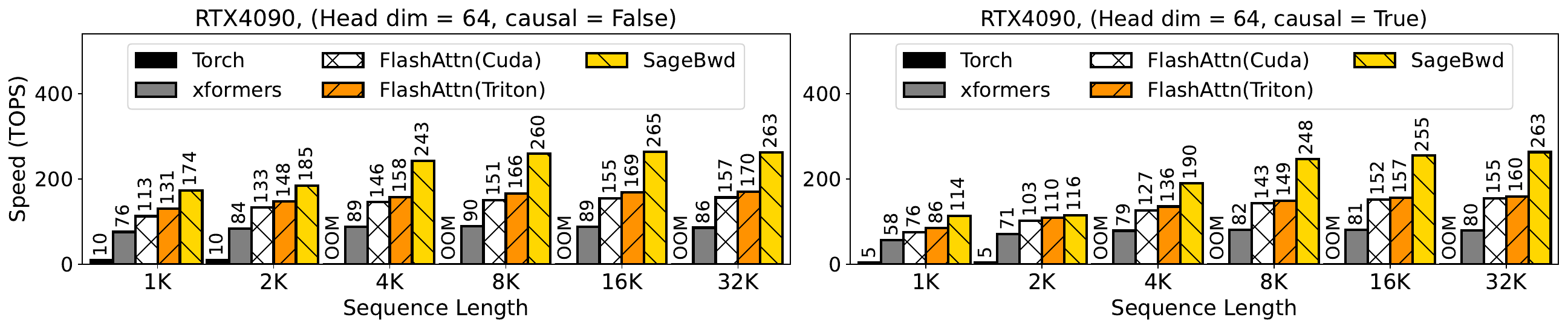}
    \vspace{-.5em}
    \caption{Speed comparison between \texttt{SageBwd} and Baselines (\texttt{RTX4090}, headim=64).}
    \label{fig:sage_train_h64_speed} 
\end{figure}

\autoref{fig:sage_train_h128_speed} and \autoref{fig:sage_train_h64_speed} report the
end-to-end forward and backward kernel throughput of \our compared to baseline
attention implementations on an \texttt{RTX4090}.
Across head dimensions $D=64$ and $D=128$, \our consistently outperforms FlashAttention2,
achieving up to a \textbf{1.67}$\times$ speedup, and exceeds the performance of
Triton- and xFormers-based FlashAttention2 implementations, too.

We note that our current implementation prioritizes correctness and stability over
aggressive kernel fusion, and further speed improvements are likely achievable with additional optimizations.

\section{Ablation Study}\label{sec:append:ablation}
In the main experiments (\autoref{sec:exp}), we apply K-smoothing by default.
However, prior SageAttention work also employs Q-smoothing as a key technique
for improving quantization accuracy~\citep{zhang2025sageattention,zhang2024sageattention2}.
In this section, we ablate the effects of Q- and K-smoothing in \our to clarify
their respective roles during pre-training.

\begin{figure}[ht]
  \centering
  \begin{subfigure}{0.48\linewidth}
    \centering
    \includegraphics[width=\linewidth]{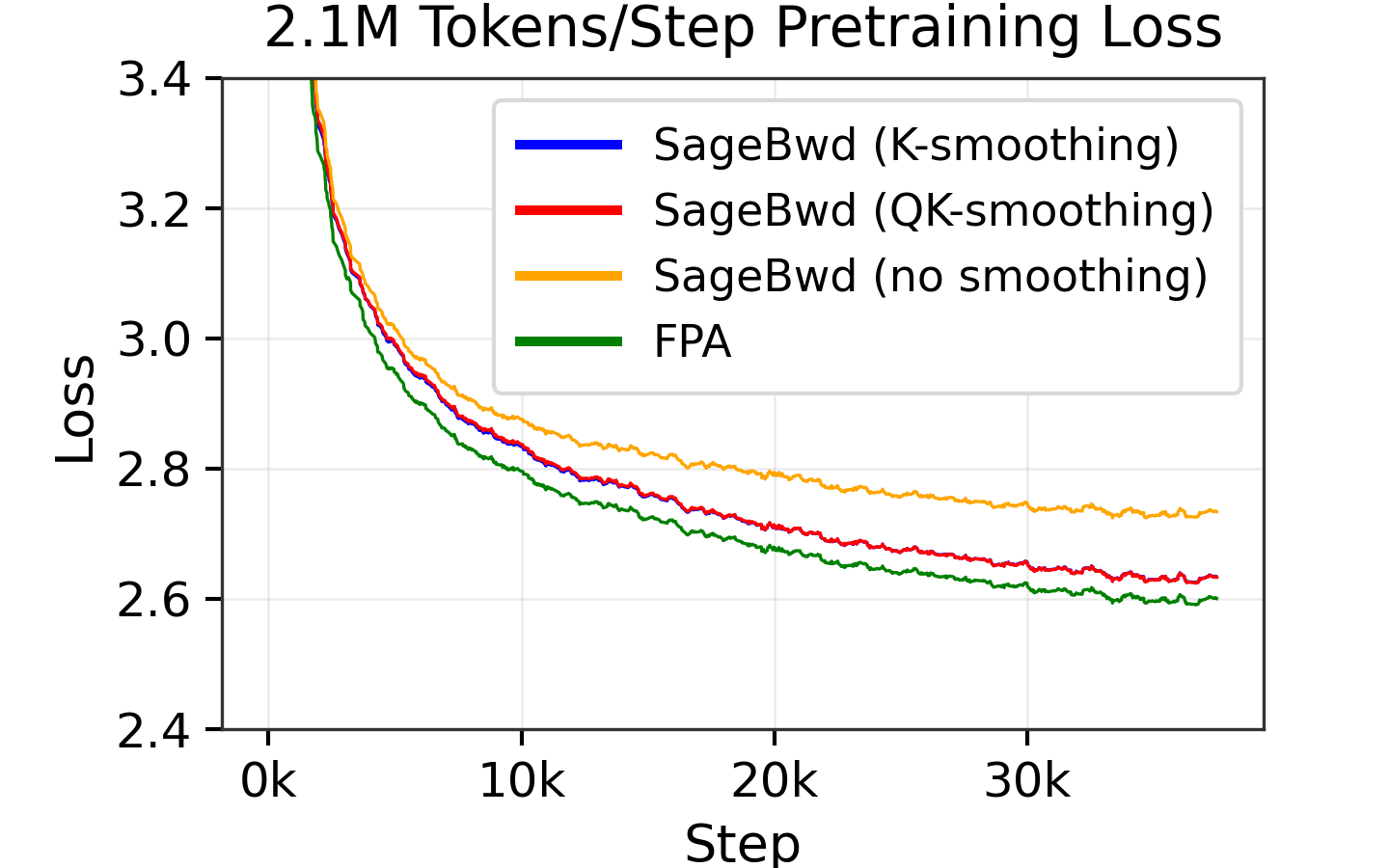}
    \caption{2.1M Tokens/Step}
    \label{fig:ablation_loss:a}
  \end{subfigure}\hfill
  \begin{subfigure}{0.48\linewidth}
    \centering
    \includegraphics[width=\linewidth]{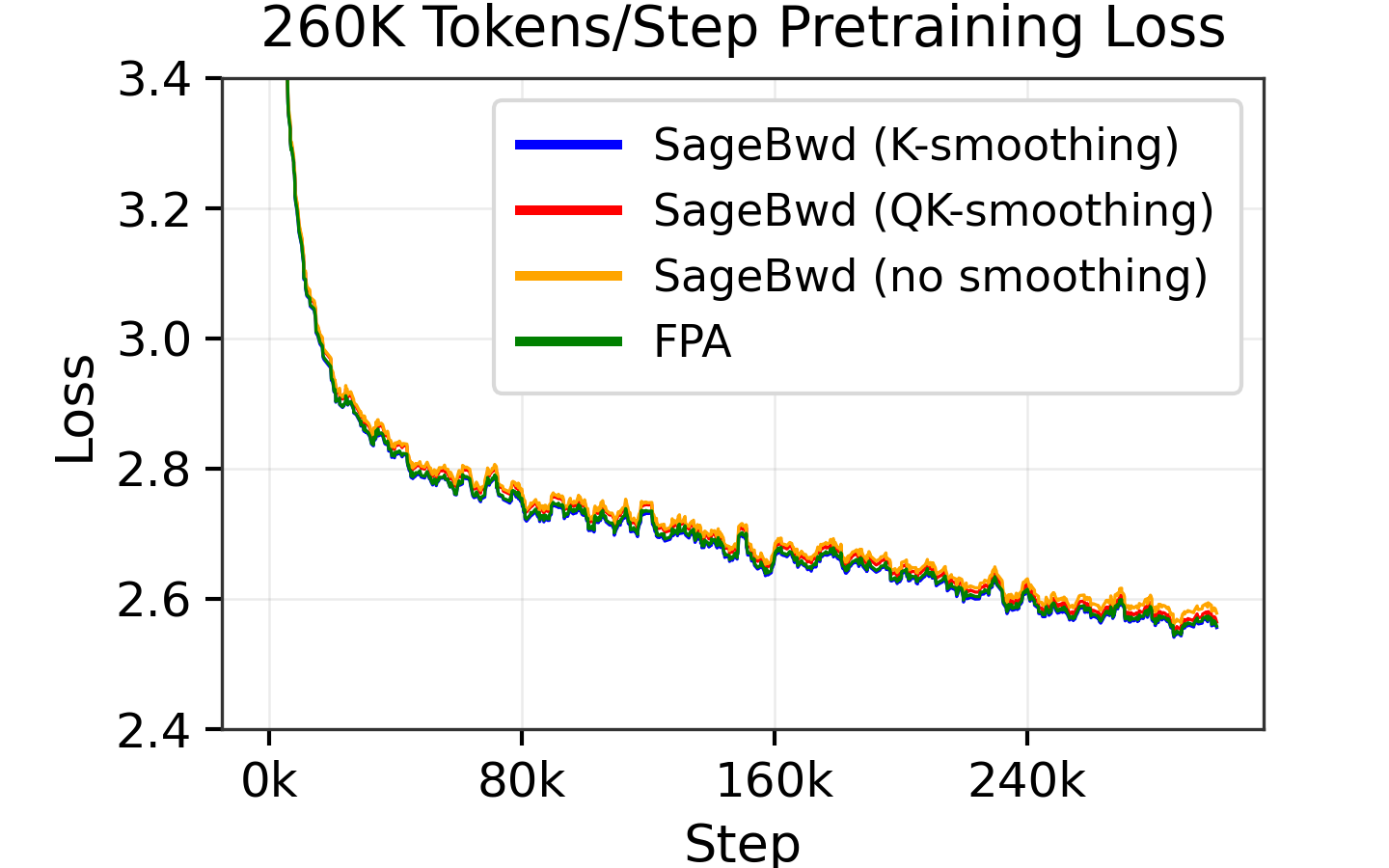}
    \caption{260K Tokens/Step}
    \label{fig:ablation_loss:b}
  \end{subfigure}
  \caption{Ablation of Q-smoothing and K-smoothing pretraining loss over 78B tokens under a different number of tokens/step}
  \label{fig:ablation_loss}
\end{figure}

In \autoref{fig:ablation_loss}, we compare full-precision attention (FPA) with
\our under three settings: no smoothing, K-smoothing, and QK-smoothing, at both
2.1M and 260K tokens per step (TPS).
Due to computational constraints, we do not evaluate Q-smoothing in isolation.
All runs use QK-norm and identical training hyperparameters to those in
\autoref{sec:exp}.

Contrary to expectations from prior work, we find that while K-smoothing remains
essential for stable pre-training, Q-smoothing provides no consistent benefit and
can slightly degrade gradient accuracy.

\paragraph{K-smoothing is necessary for stable training.}
Consistent with prior work, \autoref{fig:ablation_loss} shows that K-smoothing, a technique where the token-wise mean of $\mathbf{K}$ is subtracted prior to quantization, is critical
for maintaining pre-training stability.
Even in the more noise-tolerant 260K TPS regime, K-smoothing is required to achieve
FPA-level performance.

From an implementation perspective, K-smoothing requires no modification to the backward pass. Smoothing can occur at kernel entry and the smoothed $\mathbf{K}$ can be used without additional bias terms. 

The gradient computation for $\mathbf{dQ} = \mathbf{dS}\mathbf{K}$ remains valid even after $\mathbf{K}$ is smoothed because each row of $\mathbf{dS}$ sums to $0$, so $\mathbf{dS}(\mathbf{1} \operatorname{mean}_{\text{row}}(\mathbf{K})^\top) = 0$ and 
$\mathbf{dQ} = \mathbf{dS}\mathbf{K} = \mathbf{dS}(\mathbf{K} - \mathbf{1}\operatorname{mean}_{\text{row}}(\mathbf{K})^\top) = \mathbf{dS}\mathbf{K}^{\mathrm{sm}}$ 
where $\mathbf{K}^{\mathrm{sm}}$ denotes the key matrix $\mathbf{K}$ after smoothing.

\paragraph{Q-smoothing shows limited benefit.}
In contrast, we observe no consistent improvement in either pre-training loss or intermediate-tensor accuracy from applying Q-smoothing.
In some cases, Q-smoothing slightly degrades gradient fidelity.
As shown in our error analysis (\autoref{sec:append:cossim_l2}),
$\mathbf{dQ}$ and $\mathbf{dK}$ exhibit marginally larger deviation
from the FPA baseline when Q-smoothing is enabled.

One contributing factor is the gradient correction required by Q-smoothing.
Rewriting the logits as
$$
\mathbf{S}=(\mathbf{Q}-\mathbf{1}\mu_Q^\top)\mathbf{K}^\top + \mathbf{1}(\mu_Q\mathbf{K}^\top)
\quad\text{with}\quad
\mu_Q = \operatorname{mean}(\mathbf{Q}) .
$$
preserves forward equivalence with $\mathbf{Q}\mathbf{K}^\top$.
Therefore, the total gradient remains
$
\mathbf{dK}=\mathbf{dS}^\top \mathbf{Q}.
$
Consequently, $\mathbf{dK}$ cannot be computed from the centered branch only, i.e., $\mathbf{dK} \ne \mathbf{dK}_{\text{center}}=\mathbf{dS}^\top \mathbf{Q}^{\mathrm{sm}}$ with
$\mathbf{Q}^{\mathrm{sm}}=\operatorname{smooth}(\mathbf{Q}) = \mathbf{Q}-\mathbf{1}\mu_Q^\top$. We need to add an additional bias branch term, $\mathbf{dK}_{\text{bias}} = (\mathbf{dS}^\top \mathbf{1})\,\mu_Q^\top$, to recover the correct gradient:
$$
\mathbf{dK}=  \mathbf{dS}^\top \mathbf{Q} =  \mathbf{dS^\top}(\mathbf{Q}^{\mathrm{sm}} + \mathbf{1}\mu_Q^\top) = \mathbf{dS}^\top \mathbf{Q}^{\mathrm{sm}} + \mathbf{dS}^\top \mathbf{1}\mu_Q^\top  = \mathbf{dK}_{\text{center}}+\mathbf{dK}_{\text{bias}}.
$$

This additional correction introduces another pathway for quantization noise,
which may partially offset the benefits of reduced activation range.
We leave a deeper investigation of when Q-smoothing benefits training-time
quantization to future work.

%% file: src/6-Conclusion.tex
\section{Conclusion and Future Work}  
\paragraph{Conclusion.} In this paper, we extend \our, a trainable low-bit attention mechanism. We analyze when \our can match full-precision scaled dot-product attention during pre-training and find two key factors: (i) controlling outliers in $\mathbf Q$ and $\mathbf K$ via QK-norm is necessary for stability at large tokens-per-step, and (ii) the dominant accuracy bottleneck is the low-magnitude softmax gradient $\mathbf{dS}$, which affects $\mathbf{dQ}$ and $\mathbf{dK}$. Empirically, smaller tokens-per-step make training more tolerant to this noise, while larger tokens-per-step expose it and yield a stable but suboptimal gap.

\paragraph{Limitations and future work.}
While \our achieves FPA-level performance under moderate tokens-per-step, its training stability degrades at very large batch sizes.
A key direction for future work is therefore to develop methods that mitigate backward-pass quantization error, particularly along the $\mathbf{dS}$ path, without
relying on reduced batch size or increased gradient noise. In addition, although \our already delivers considerable speedups over existing baselines, further kernel-level optimizations remain an important avenue for future research.

%% file: src/Appendix.tex
\appendix
\section{\our Algorithm}\label{sec:append:algorithm}
\subsection{Forward Pass}
\begin{algorithm}[ht]
    \small
    \caption{Foward pass of the \texttt{8-bit} attention.}
    \label{alg:int8_train_fwd} 
    \begin{algorithmic}[1]
    \STATE {\bfseries Input:} {\texttt{FP16} matrices $Q, K, V \in \mathbb{R}^{N \times d}$, and block size $B_q, B_{kv}$.}
    
    \STATE Divide $Q$ to $T_m = {N}/{B_q}$ blocks $\{\vQ_i\}$; divide $K$, and $V$ to $T_n = {N}/{B_{kv}}$ blocks $\{\vK_i\}$, $\{\vV_i\}$ ;  

    \STATE {\bfseries Quantization:} \jt{$\{\mathbf{s_Q}, \hat \vQ_i\} = \{\psi (\vQ_i)\}$, ~~ $\{\mathbf{s_K}, \hat \vK_i\} = \{\psi (\vK_i^\top)\}$, ~~$\{\mathbf{s_V}, \hat \vV_i\} = \{\psi (\vV_i)\}$} ; \annotate{// Per-block.}

    \FOR {$i=1$ {\bfseries to} $T_m$}

    \STATE $\vO_i \in \mathbb{R}^{B_q\times D}= (0)$, ~~ $\mathbf{L}_i \in \mathbb{R}^{B_q} = (0),~~ m_i \in \mathbb{R}^{B_{kv}} = (0)$ ;

        \FOR {$j$ in [1, $T_n$]} 

            \STATE \jt{$\vS_{ij} = \texttt{MM} (\hat \vQ_i, \hat \vK_j) \times \mathbf{s_Q} \times \mathbf{s_K}$} ;  

            \STATE $m_{ij} = \mathrm{max}(m_{i,j-1}, \mathrm{rowmax}(\vS_{ij}))$, $\widetilde \vP_{ij} = \mathrm{exp}(\vS_{ij} - m_{ij})$, $l_{ij} = e^{m_{i,j-1}-m_{ij}} + \mathrm{rowsum}(\widetilde \vP_{ij})$;

            \STATE \jt{$\mathbf{s_P} = \mathrm{exp}(\mathrm{rowmax}(\vS_{ij}) - m_{ij}) / 127$, ~~ $\mathbf{\hat \vP}_{ij} = \widetilde \vP_{ij} / \mathbf{s_P}$} ; \annotate{// Per-token quantization.}

            \STATE $\vO_{ij} = \mathrm{diag}(e^{m_{i,j-1}-m_{ij}})^{-1} \vO_{i,j-1} + \jt{\texttt{MM}(\hat \vP_{ij}, \hat \vV_j) \times \mathbf{s_{P}} \times \mathbf{s_V}}$
        
        \ENDFOR
        \STATE $\vO_i = \mathrm{diag}(l_{i,T_n}) ^{-1} \vO_{i,T_n}$ ; 
        \STATE $\mathbf{L}_i = m_{i, T_n} + \mathrm{log}(l_{i, T_n})$ ; 
    \ENDFOR
    
    \STATE \textbf{return} $O = \{\vO_i\}$, $L = \{\mathbf{L}_i\}$ ;
    \end{algorithmic}
    \vspace{-.15em}
\end{algorithm}

\subsection{Backward Pass}
\begin{algorithm}[ht]
    \small
    \caption{Backward pass of the \texttt{8-bit} attention.}
    \label{alg:int8_train_bwd} 
    \begin{algorithmic}[1]
    \STATE {\bfseries Input:} \jt{$\{\mathbf{s_Q}, \hat \vQ_i\}, \{\mathbf{s_K}, \hat \vK_i\}$, $\{\mathbf{s_V}, \hat \vV_i\}$}, $O$, $\{\mathbf{L}_i\}$ from the forward, $dO \in \mathbb{R}^{N \times d}$, and block size $B_q, B_{kv}$ ; 

    \STATE $D = \mathrm{rowsum}(dO \circ O)$, ~~ divide $D$ to $T_m = {N}/{B_q}$ blocks $\{\mathbf{D}_i\}$; 

    \FOR {$j=1$ {\bfseries to} $T_n$}

        \FOR {$i$ in [1, $T_m$]} 

            \STATE \jt{$\vS_{ij} = \texttt{MM} (\hat \vQ_i, \hat \vK_j) \times \mathbf{s_Q} \times \mathbf{s_K}$} ;  ~~~~~$\vP_{ij} = \mathrm{exp}(\vS_{ij} - \mathbf{L}_{i})$ ;

            \STATE \jt{$\mathbf{s_P}, \hat \vP_{ij} = \psi(\vP_{ij})$, ~~ $\mathbf{s_{dO}}, \hat \vdO_{i} = \psi(\vdO_{i})$} ; ~\annotate{// \texttt{INT8} per-block quantization.}

            \STATE $\vdV_j \leftarrow \vdV_j + \jt{\texttt{MM}(\hat \vP_{ij}^\top, \hat \vdO_{i}) \times \mathbf{s_P} \times \mathbf{s_{dO}}}$ ;

            \STATE $\vdP_{ij} = \texttt{MM}(\vdO, \vV_j^\top)$ ;  ~\annotate{// Keep in \texttt{FP16}.}

            \STATE $\vdS_{ij} = \vP_{ij} \circ (\vdP_{ij} - \mathbf{D}_i)$ ; ~~~~~\jt{$\mathbf{s_{dS}}, \hat \vdS_{ij} = \psi(\vdS_{ij})$} ; ~\annotate{// \texttt{INT8} per-block quantization.}

            \STATE $\vdQ_i \leftarrow \vdQ_i + \jt{\texttt{MM}(\hat \vdS_{ij}, \hat \vK_{j}) \times \mathbf{s_{dS}} \times \mathbf{s_K}}$ ;

            \STATE $\vdK_j \leftarrow \vdK_j + \jt{\texttt{MM}(\hat {\vdS}^\top_{ij}, \hat \vQ_{i}) \times \mathbf{s_{dS}} \times \mathbf{s_Q}}$ ;

        \ENDFOR

    \ENDFOR
    
    \STATE \textbf{return} $dQ, dK, dV$ ;
    \end{algorithmic}
    \vspace{-.15em}
\end{algorithm}

\newpage

\section{dS Magnitude}\label{sec:append:ds}
In this section, we prove a simple upper bound on the RMS of $\mathbf{dS}$.

\paragraph{Proof.}
Recall that
$$
\mathbf{dS} = \mathbf{P} \circ (\mathbf{dP} - \boldsymbol{\delta}\mathbf{1}^\top) \in \mathbb{R}^{N \times N}, 
\qquad 
\boldsymbol{\delta} = \operatorname{rowsum}(\mathbf{dO} \circ \mathbf{O}) \in \mathbb{R}^{N},
$$
where $\circ$ denotes elementwise multiplication and $\mathbf{1}$ is the all-ones vector.

For row $i$ of $\mathbf{dS}$, we can write
$$
\mathbf{dS}_i = \mathbf{P}_i \circ (\mathbf{dP}_i - \boldsymbol{\delta}_i\mathbf{1}),
$$
where $\mathbf{P}_i, \mathbf{dP}_i \in \mathbb{R}^N$ and $\boldsymbol{\delta}_i$ is the $i$-th entry of $\boldsymbol{\delta}$.
Its root-mean-square (RMS) value is
$$
\operatorname{RMS}(\mathbf{dS}_i)
= \sqrt{\frac{1}{N}\sum_{j=1}^N \mathbf{P}_{i,j}^2 (\mathbf{dP}_{i,j} - \boldsymbol{\delta}_i)^2 }.
$$

Using the infinity norm ($\|\mathbf{x}\|_\infty = \max_j |\mathbf{x}_j|$), we have $|\mathbf{dP}_{i,j} - \boldsymbol{\delta}_i| \;\le\; \|\mathbf{dP}_i - \boldsymbol{\delta}_i\mathbf{1}\|_\infty$. Therefore,
\begin{align}
\operatorname{RMS}(\mathbf{dS}_i)^2
&\le \|\mathbf{dP}_i - \boldsymbol{\delta}_i\mathbf{1}\|_\infty^2 \cdot \frac{1}{N}\sum_{j=1}^N \mathbf{P}_{i,j}^2 \\
&= \|\mathbf{dP}_i - \boldsymbol{\delta}_i\mathbf{1}\|_\infty^2 \, \operatorname{RMS}(\mathbf{P}_i)^2,
\end{align}
and hence
\begin{equation}
\operatorname{RMS}(\mathbf{dS}_i)
\le \operatorname{RMS}(\mathbf{P}_i) \, \|\mathbf{dP}_i - \boldsymbol{\delta}_i\mathbf{1}\|_\infty.
\label{eq:ds-bound-p}
\end{equation}

Because $\mathbf{P}$ is the output of a softmax operation, each row $\mathbf{P}_i$ is a probability vector that sums to $1$ and only has entries in the range $[0,1]$. Thus
\begin{equation}
\operatorname{RMS}(\mathbf{P}_i)
= \sqrt{\frac{1}{N}\sum_{j=1}^N \mathbf{P}_{i,j}^2}
\;\le\; \sqrt{\frac{1}{N}\max_j \mathbf{P}_{i,j} \sum_{j=1}^N \mathbf{P}_{i,j}}
\;\le\; \frac{1}{\sqrt{N}}.
\label{eq:p-rms-bound}
\end{equation}

Combining \eqref{eq:ds-bound-p} and \eqref{eq:p-rms-bound} yields, for each row $i$,
$$
\operatorname{RMS}(\mathbf{dS}_i)
\le \frac{1}{\sqrt{N}} \, \|\mathbf{dP}_i - \boldsymbol{\delta}_i\mathbf{1}\|_\infty.
$$

Finally, the global RMS of $\mathbf{dS}$ satisfies
\begin{equation}
\operatorname{RMS}(\mathbf{dS})
\le \frac{1}{\sqrt{N}} \, \max_{i} \| \mathbf{dP}_i - \boldsymbol{\delta}_i\mathbf{1}\|_\infty.
\label{eq:ds-global-bound}
\end{equation}
This upper bound can be interpreted as follows: the average magnitude of $\mathbf{dS}$ is at most
the largest per-row gradient magnitude in $\mathbf{dP}$, scaled by a factor of $1/\sqrt{N}$.

\newpage

\section{Cosine Similarity and Rel-L2 Error}\label{sec:append:cossim_l2}

\begin{figure}[h!]
  \centering
  \begin{subfigure}{0.48\linewidth}
    \centering
    \includegraphics[width=\linewidth]{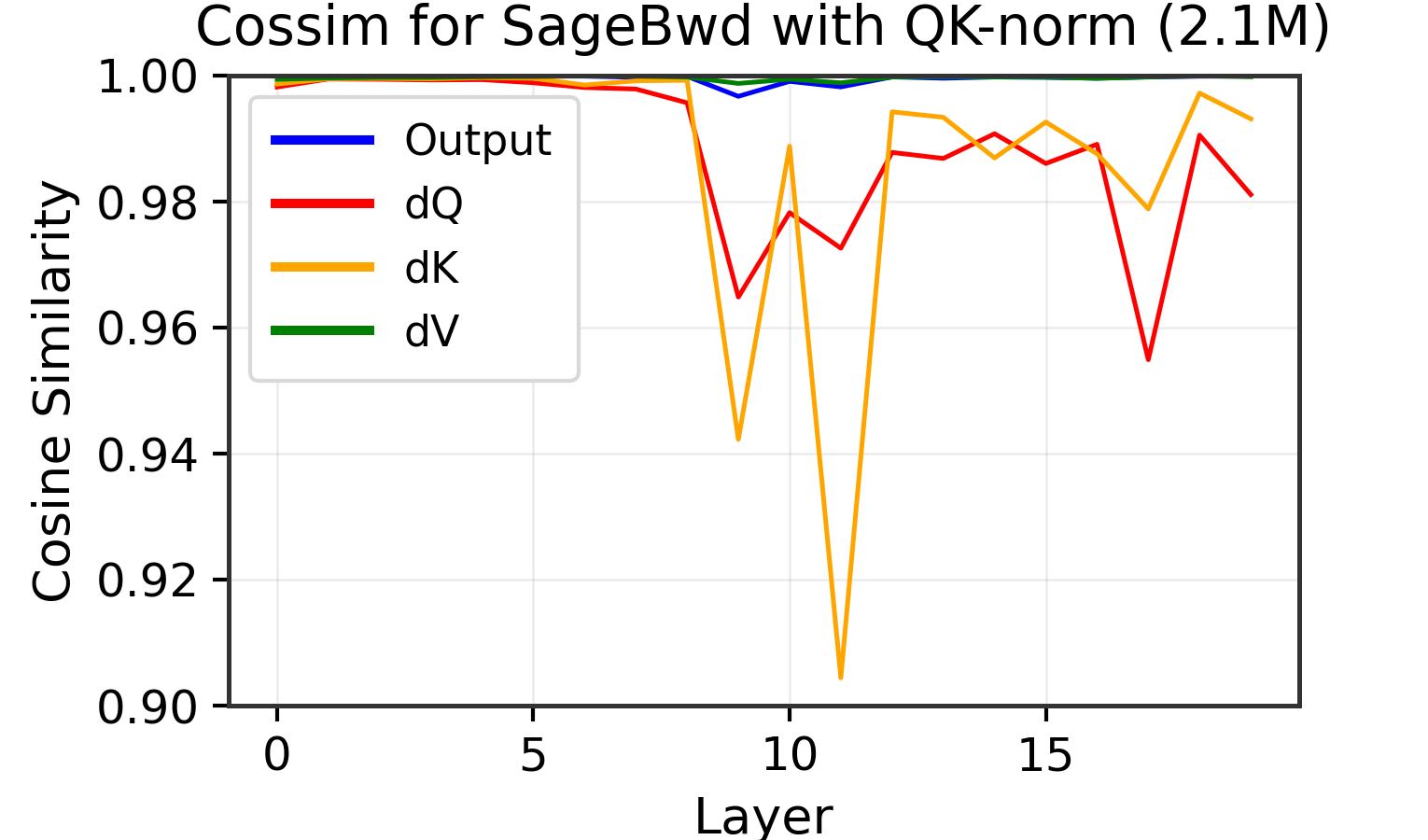}
    \caption{}
    \label{fig:metrics:cossim:a}
  \end{subfigure}\hfill
  \begin{subfigure}{0.48\linewidth}
    \centering
    \includegraphics[width=\linewidth]{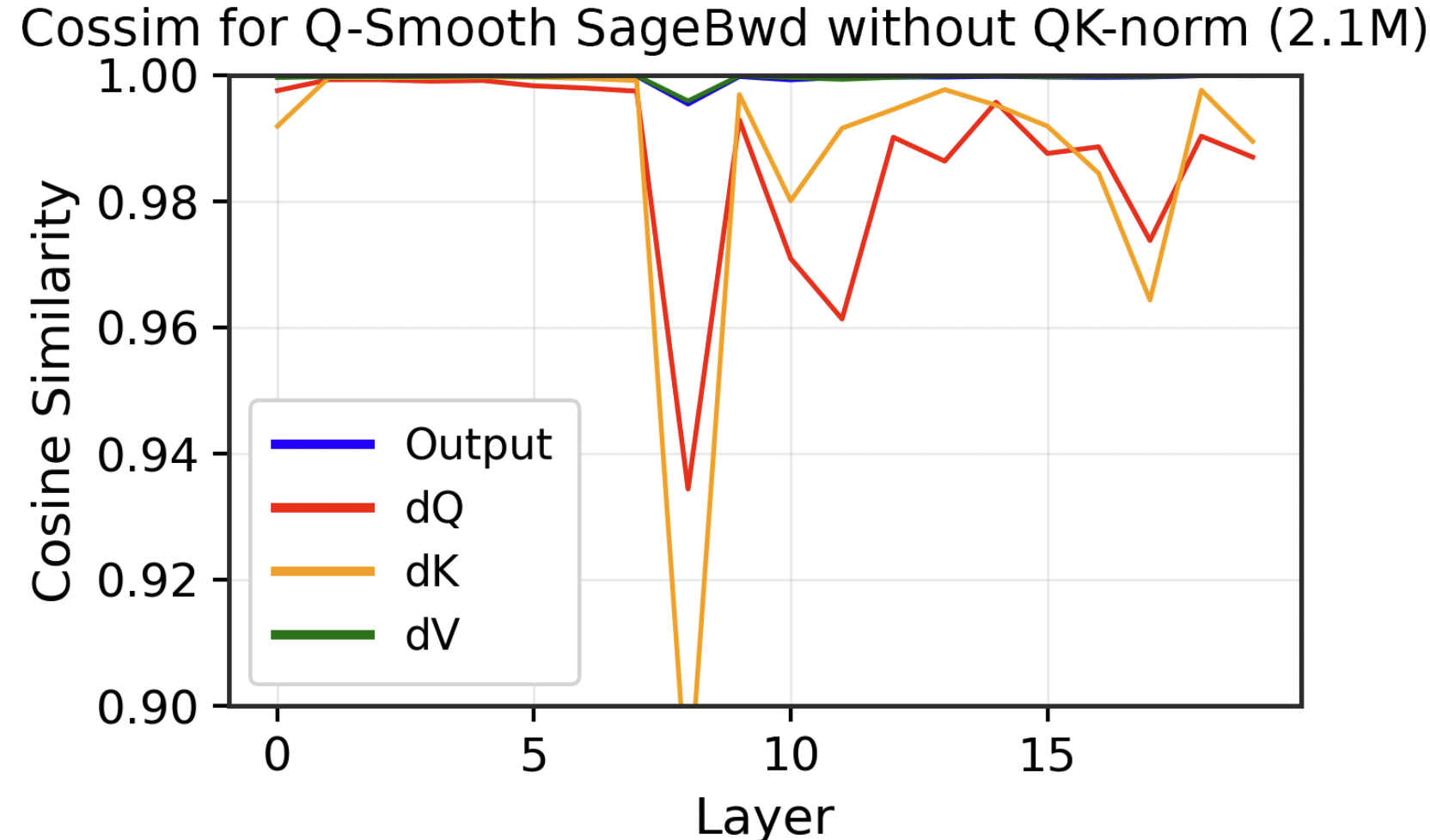}
    \caption{}
    \label{fig:metrics:cossim:b}
  \end{subfigure}
  \begin{subfigure}{0.48\linewidth}
    \centering
    \includegraphics[width=\linewidth]{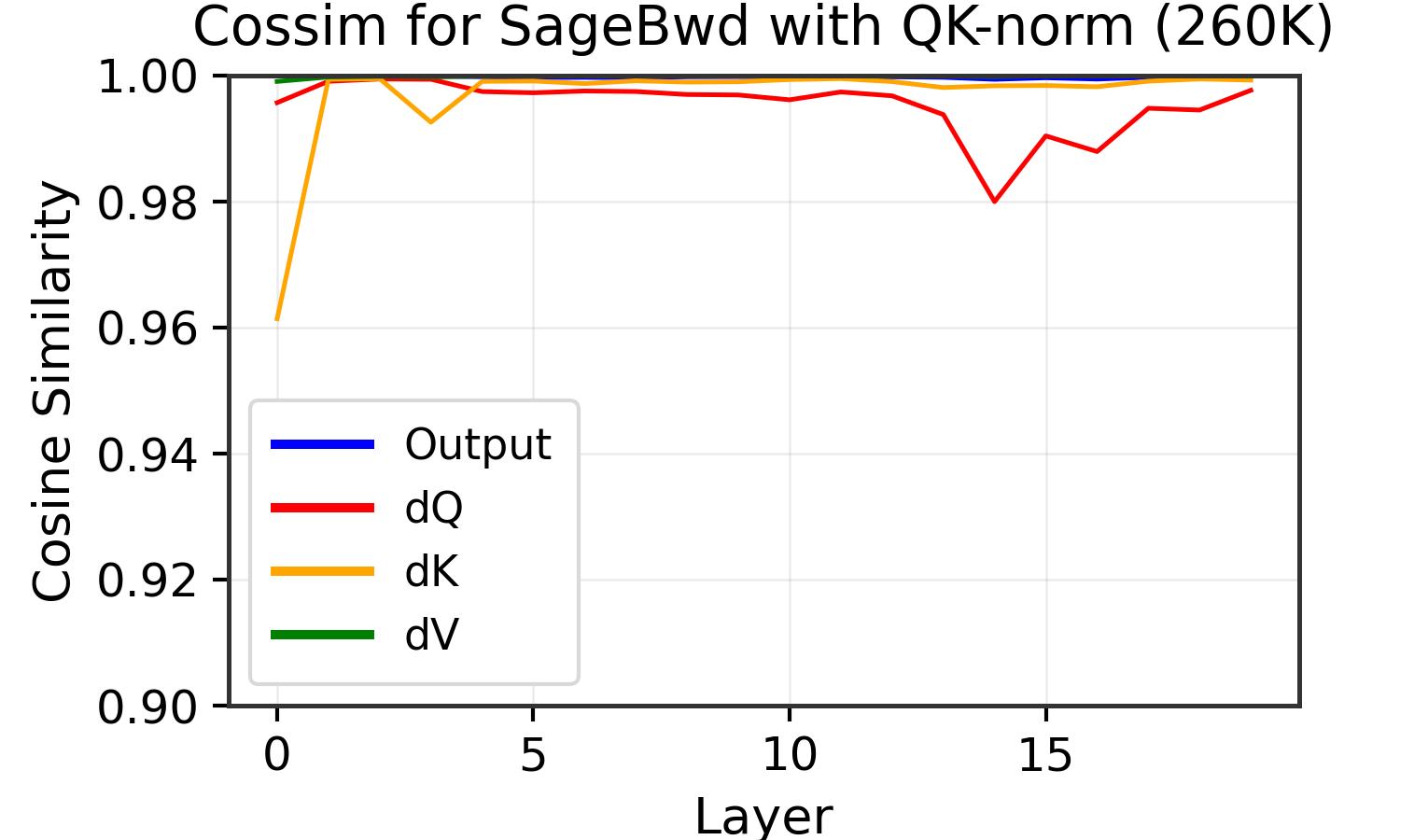}
    \caption{}
    \label{fig:metrics:cossim:c}
  \end{subfigure}
  \begin{subfigure}{0.48\linewidth}
    \centering
    \includegraphics[width=\linewidth]{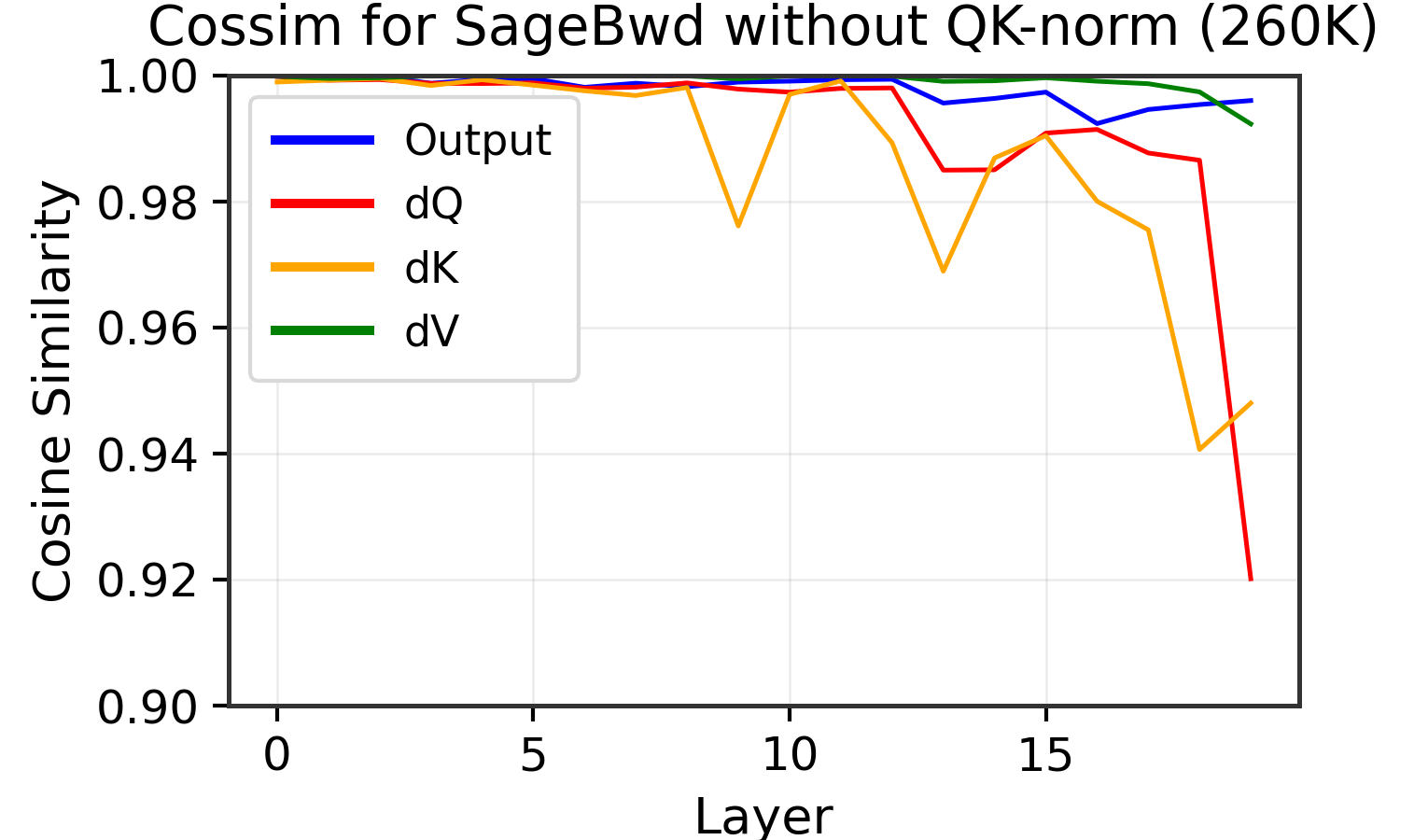}
    \caption{}
    \label{fig:metrics:cossim:d}
  \end{subfigure}
  \caption{Cosine similarity between \our and SDPA over layers on different settings}
  \label{fig:metrics:cossim}
\end{figure}

\begin{figure}[h!]
  \centering
  \begin{subfigure}{0.48\linewidth}
    \centering
    \includegraphics[width=\linewidth]{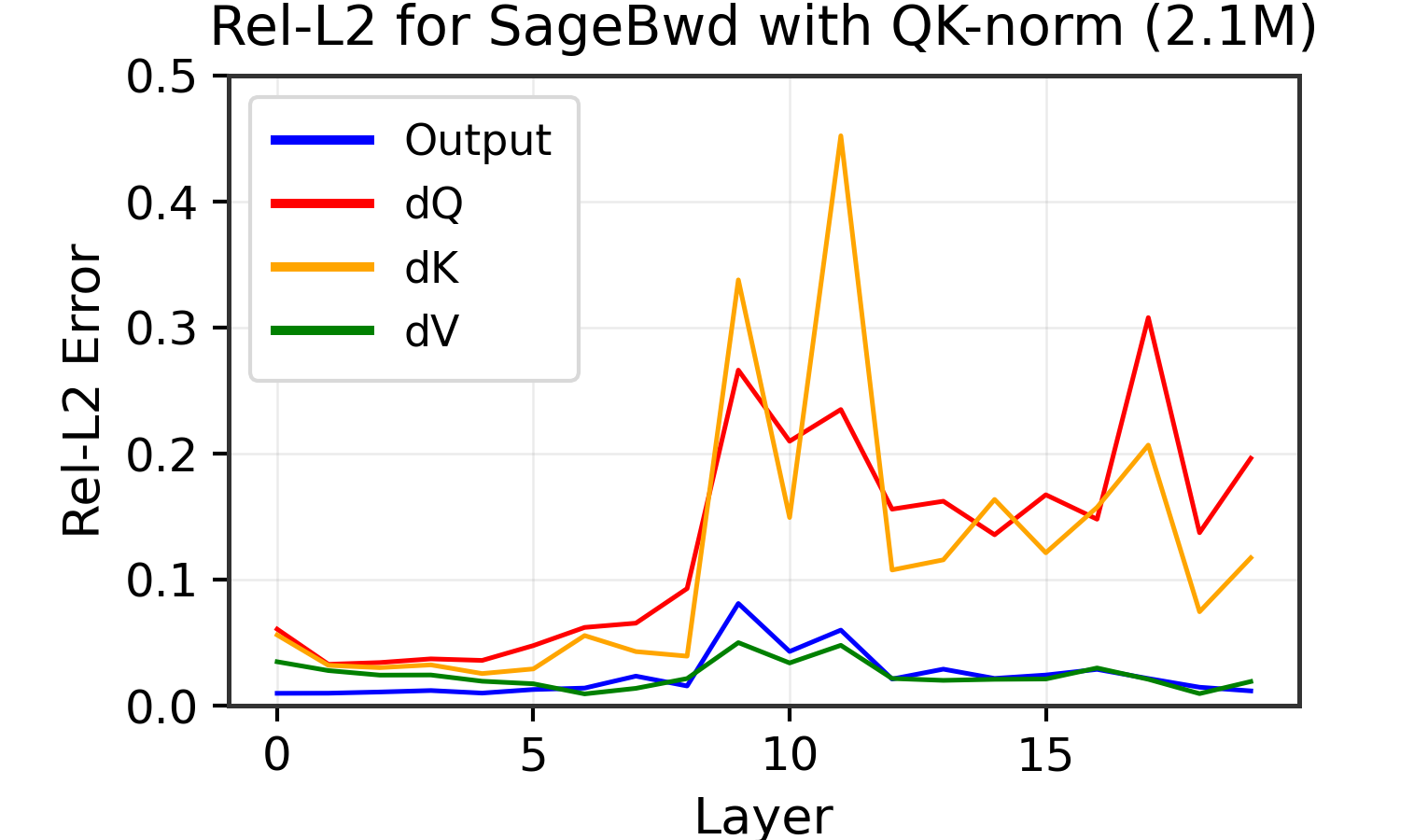}
    \caption{}
    \label{fig:metrics:l2:a}
  \end{subfigure}\hfill
  \begin{subfigure}{0.48\linewidth}
    \centering
    \includegraphics[width=\linewidth]{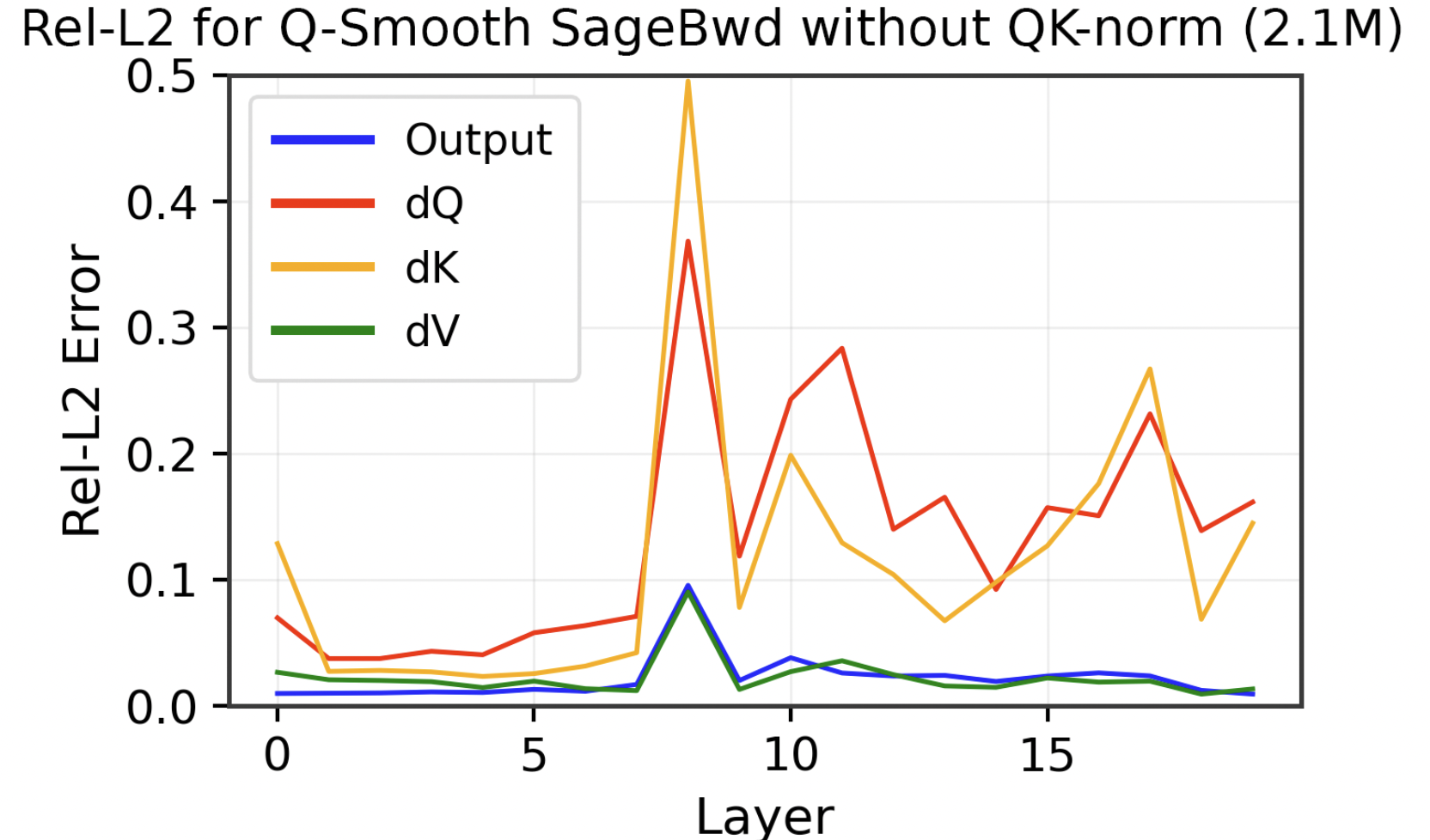}
    \caption{}
    \label{fig:metrics:l2:b}
  \end{subfigure}
  \begin{subfigure}{0.48\linewidth}
    \centering
    \includegraphics[width=\linewidth]{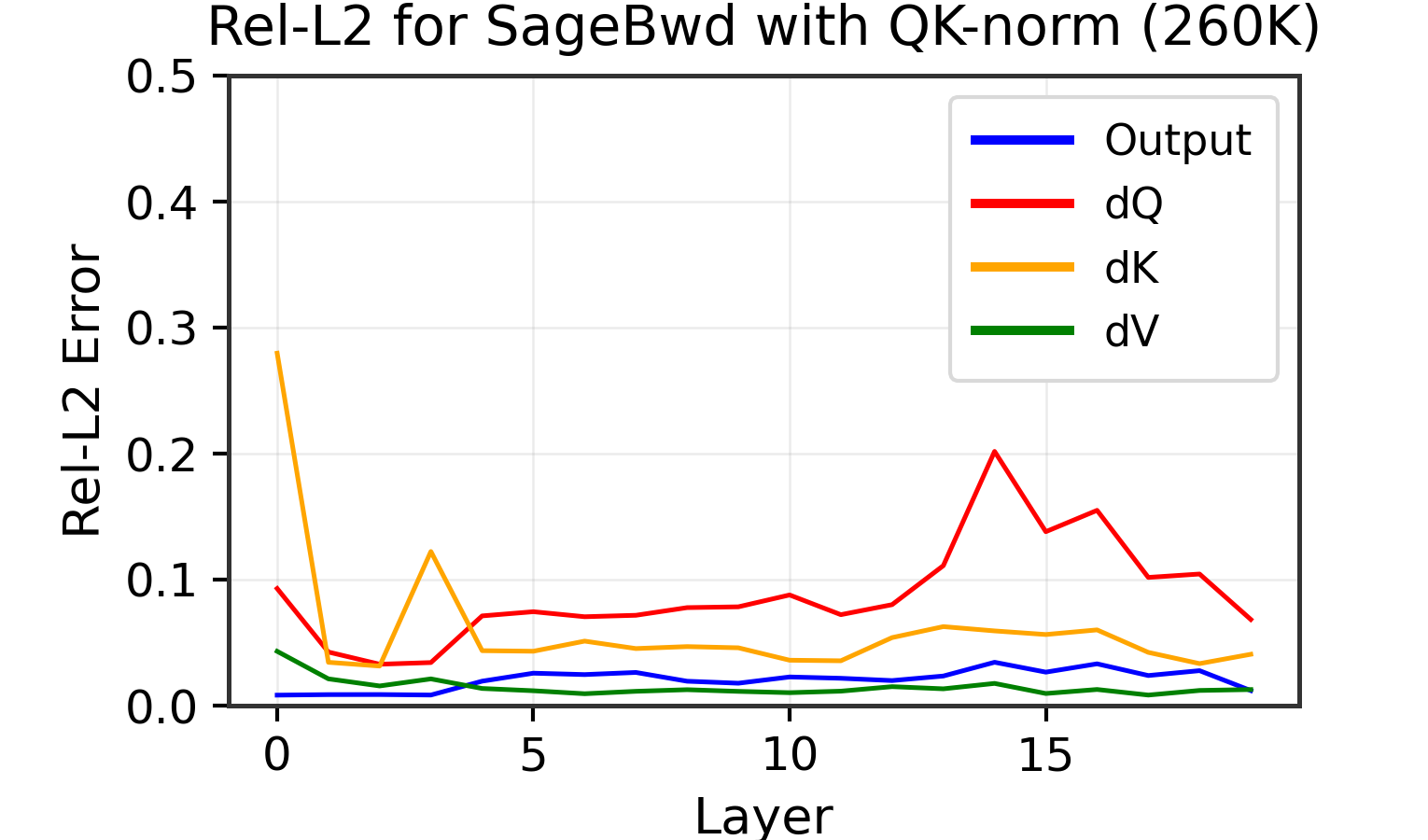}
    \caption{}
    \label{fig:metrics:l2:c}
  \end{subfigure}
  \begin{subfigure}{0.48\linewidth}
    \centering
    \includegraphics[width=\linewidth]{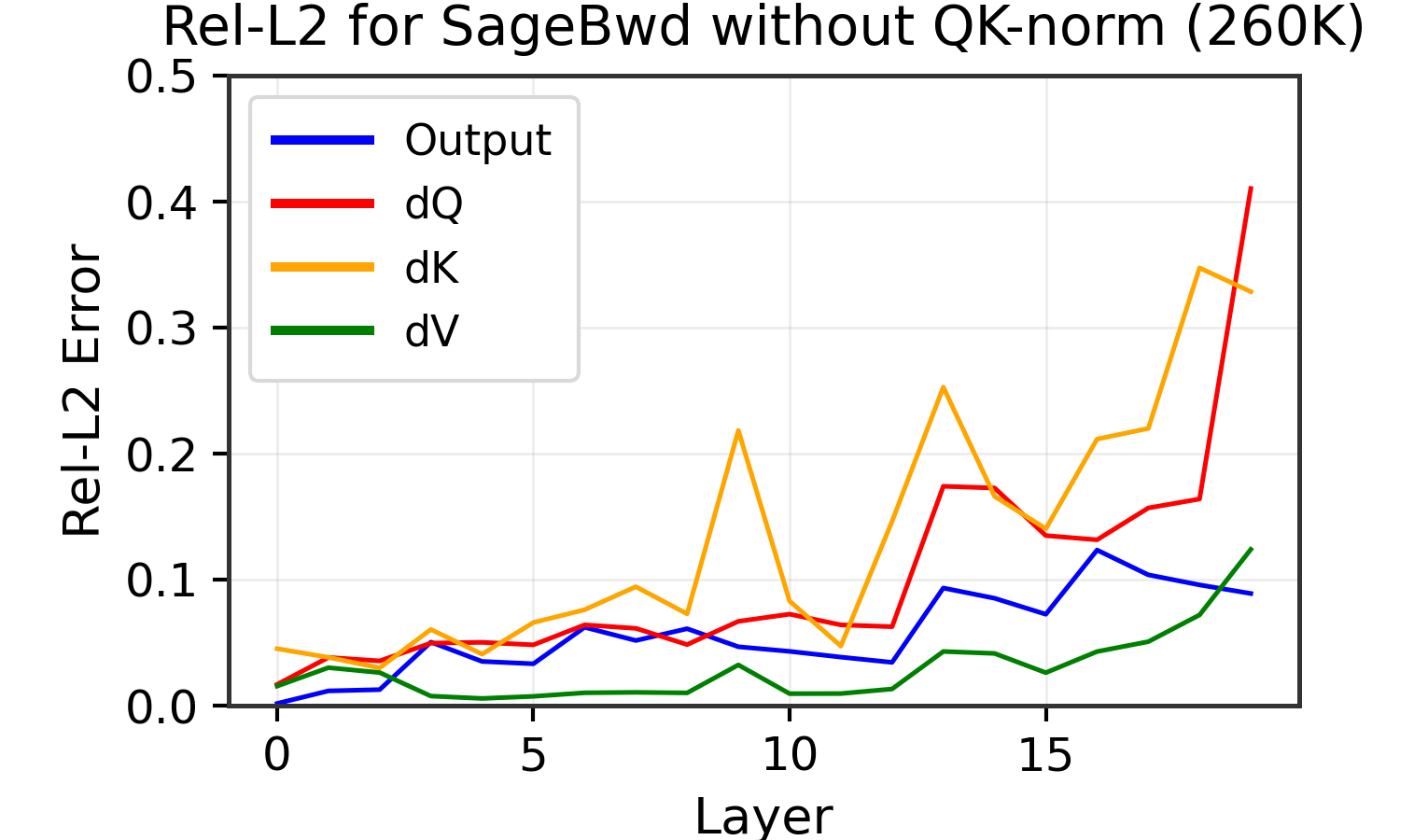}
    \caption{}
    \label{fig:metrics:l2:d}
  \end{subfigure}
  \caption{Relative L2-Error between \our and SDPA over layers on different settings}
  \label{fig:metrics:l2}
\end{figure}

\autoref{fig:metrics:cossim} and \autoref{fig:metrics:l2} show the cosine similarity and relative $\ell_2$-error between \our and FPA on inputs and gradients extracted from a single forward-backward pass of the pretrained 325M Llama model under various TPS and architectural settings.

